# Aggregated Pyramid Vision Transformer: Split-transform-merge Strategy for Image Recognition without Convolutions

Rui-Yang Ju
Department of Electrical and
Computer Engineering
Tamkang University
jryjry1094791442@gmail.com

Ting-Yu Lin
Department of Engineering
Science
National Cheng Kung University
tonylin0413@gmail.com

Jen-Shiun Chiang
Department of Electrical and
Computer Engineering
Tamkang University
jsken.chiang@gmail.com

Jia-Hao Jian
Department of Electrical and
Computer Engineering
Tamkang University
jianjiahao408@gmail.com

Yu-Shian Lin
Department of Electrical and
Computer Engineering
Tamkang University
abcpp12383@gmail.com

Liu-Rui-Yi Huang
Department of Electrical and
Computer Engineering
Tamkang University
ruiyihuangliu@gmail.com

*Abstract*—With the achievements of Transformer in the field of natural language processing, the encoder-decoder and the attention mechanism in Transformer have been applied to computer vision. Recently, in multiple tasks of computer vision (image classification, object detection, semantic segmentation, etc.), state-of-the-art convolutional neural networks have introduced some concepts of Transformer. This proves that Transformer has a good prospect in the field of image recognition. After Vision Transformer was proposed, more and more works began to use self-attention to completely replace the convolutional layer. This work is based on Vision Transformer, combined with the pyramid architecture, using Split-transform-merge to propose the group encoder and name the network architecture Aggregated Pyramid Vision Transformer (APVT). We perform image classification tasks on the CIFAR-10 dataset and object detection tasks on the COCO 2017 dataset. Compared with other network architectures that use Transformer as the backbone, APVT has excellent results while reducing the computational cost. We hope this improved strategy can provide a reference for future Transformer research in computer vision.

## I. INTRODUCTION

Inspired by the self-attention mechanism [1-2], the proposed Transformer [3] has become the most important model architecture for natural language processing (NLP). Devlin [4] fine-tuned the Transformer to complete tasks on small task-specific datasets. The research of Brown [5] and Lepikhin [6] proved that Transformer can realize the train models of big size, which makes it have more applications.

In the past few years, Convolutional Neural Networks (CNN) have been widely used in the field of Computer Vision (CV), and multiple state-of-the-art network architectures [7-10] have achieved success in dense prediction tasks.

With the success of Transformer in NLP, multiple works [11-16] began to add self-attention to CNN. Cordonnier [17] studied the direct relationship between self-attention and convolution layers, proving that the two can be integrated or replaced with each other. As self-attention plays a better role in CNN, researchers began to consider abandoning the convolutional layer as the primary block. Experiments [18-19] show that self-attention can completely replace the convolutional layer, and it achieves better results in the task of CV. Dosovitskiy [20] first applied the entire Transformer architecture to CV tasks, although Vision Transformer (ViT) did not perform well on small-size datasets, it made more researchers start to consider Transformer as the primary block. Zhai [21] improved the ViT architecture to solve the problem of Transformer scaling, further improving the accuracy of ViT in ImageNet. Fan [22] proposed MViT with ViT as the basic architecture, and achieved the state-of-the-art result in image classification tasks.

In recent years, Transformer backbones have been widely used in image classification tasks. DeiT [23] introduced a new training strategy in ViT and proposed token-based distillation to improve experimental results. T2T-ViT [24] proposed a layer-by-layer transformation from tokens to tokens to reduce the parameters of the model. Patch embedding is an important part of Transformer backbones research. Han [25] used the internal and external Transformer blocks to generate pixel embedding and patch embedding respectively, and proposed the Transformer iN Transformer (TNT) model to improve accuracy. Chu [26] also studied the patch of ViT, using the Position Encoding Generator (PEG) to replace the fixed size position in ViT, making the model easier to deal with images of different resolutions. However, Chen [27] used a dual-branch Transformer to patch images of different sizes. Deep convolution has achieved good results in the application of CNN network architecture, and Local-ViT [28] fused it into ViT to improve the local continuity of features.

In the process of dealing with dense prediction tasks (object detection, semantic segmentation, etc.), Wang [29-30] realized that ViT has the disadvantages of excessive computational cost and low resolution of output feature maps, so they proposed the Pyramid Vision Transformer (PVT), which uses pyramids model to overcome these difficulties. Swin Transformer [31-32] replaced fixed-size positional embeddings with relative positional deviations, defeating many CNN architectures, once again proving that Transformer is very promising in the field of computer vision. While CvT [34], CoaT [35] and LeViT [36] improved the accuracy by introducing convolution-like operations into ViT.

Obviously, researchers have conducted in-depth research on patch, attention mechanism and other parts of Transformer, and have also applied the operation of convolution to Transformer, but there are few works on improving the overall architecture of Transformer. Inspired by the split-transform-merge of Inception [37], we apply the group convolution in ResNeXt [38] model to the Transformer and propose the group encoder. We use aggregated Transformer to replace the entire Transformer architecture in ViT, obtaining more features through group encoders to improve the experimental results.

There are two main contributions of this paper.

1) We propose Aggregated Pyramid Vision Transformer (APVT), which uses the group encoder in APVT to obtain more features, and all our paths share the same topology in each aggregated Transformer.

2) We emphasize that the experimental results can be improved through the improvement of patch, attention mechanism and other parts of Transformer, but the literatures on improving the entire Transformer architecture are rare. Our approach shows that the number of branches of the encoder is a concrete, measurable dimension that is of central importance. Experiments show that increasing the number of branches of the encoder is a more effective way to obtain accuracy.

## II. RELATED WORKS

### A. Transformer Backbones

Vaswani [3] first proposed Transformer and applied it to machine translation tasks, becoming the preferred method for task research in NLP. But the attention mechanism cannot be directly applied to the image because it would incur extra cost if every pixel of the image pays attention to each other. Local multi-head dot-product self-attention [39-41] solves this problem, discarding the convolutional layer in the network architecture.

Cordonnier [17] obtains features from input images by extracting patches and applying the attention mechanism. ViT [20] adopts the same method to extract 2 × 2 pixels patches from Height × Width (H × W) input images. The number of patches is $H \times W / P^2$, and reshape these patches into one-dimensional data, as input to Transformer. The encoder consists of multiheaded self-attention (MSA) and MLP blocks. The formula of MSA is as follows:

$$MSA(z) = [SA_1(z); SA_2(z); \cdots SA_k(z)]U_{msa} \quad (1)$$

where SA is Standard qkv self-attention [3], and for each element in the input sequence $z \in \mathbb{R}^{N \times D}$, MSA runs the self-attention of k heads in parallel.

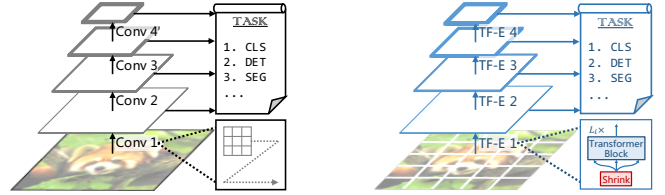

Fig. 1. **Left**: The pyramid architecture in CNN. **Right**: The pyramid architecture in Transformer. [29]

### B. Pyramid Architecture

In the field of computer vision, reasonable extraction of features of different scales is a difficult problem to overcome. The traditional image pyramid is not very efficient in terms of time and space costs. Feature pyramid [42] combined the encoder-decoder architecture to solve this problem. As shown in Fig. 1, the pyramid architecture in Transformer is similar to the architecture in CNN [43], using 4 stages to obtain features of different scales. But unlike CNN using different convolutions to obtain feature map, Transformer obtains features through patch embedding. Different from the 2 × 2 pixels patch used by the traditional Transformer, pyramid architecture uses a 4 × 4 pixel patch for high-resolution input images. In order to reduce computational cost, it reduces the sequence length of Transformer as the network deepens. To reduce the MSA [20] used by Transformer, the spatial-reduction attention (SRA) [29] formula is proposed as follows:

$$SRA(Q, K, V) = Concat(head_0, \cdots head_N)W^O \quad (2)$$

$$head_j = Attention(QW_j^Q, SR(K)W_j^K, SR(V)W_j^V) \quad (3)$$

$$SR(x) = Norm(Reshape(x, R_i)W^S) \quad (4)$$

$$Attention(q, k, v) = Softmax(\frac{qk^T}{\sqrt{d_{head}}})V \quad (5)$$

where Concat (·) is the concatenation operation[3], Eq. (3) is the operation to reduce the spatial dimension of the input sequence, and Eq. (4) is the attention operation.

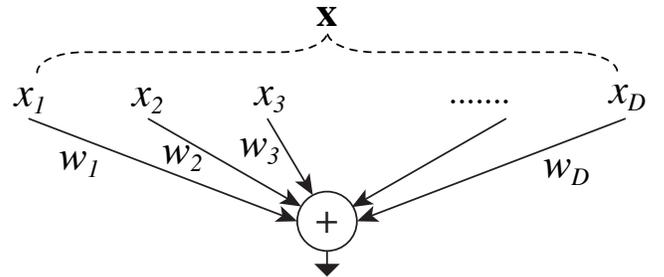

Fig. 2. Split-transform-merge: the neuron that performs inner product. [38]

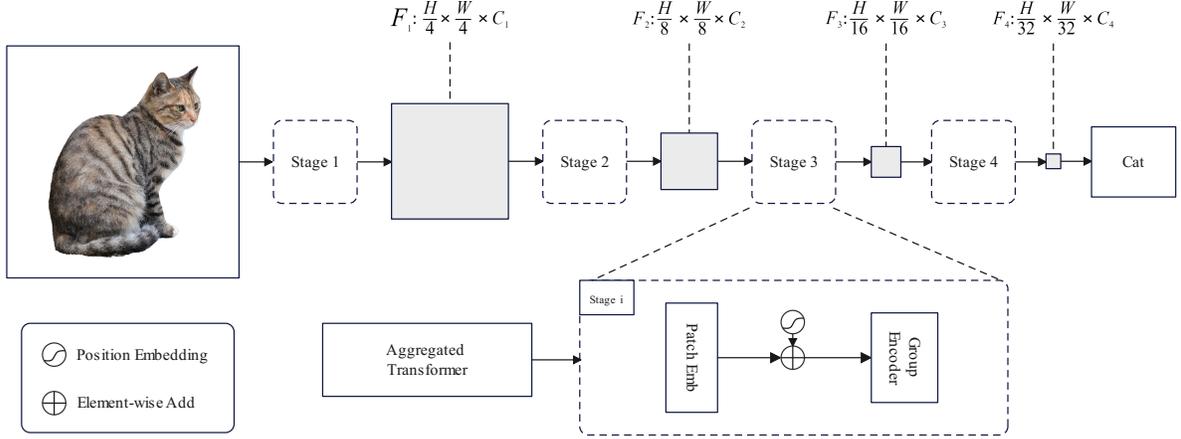

Fig. 3. The overall architecture of Aggregated Pyramid Vision Transformer.

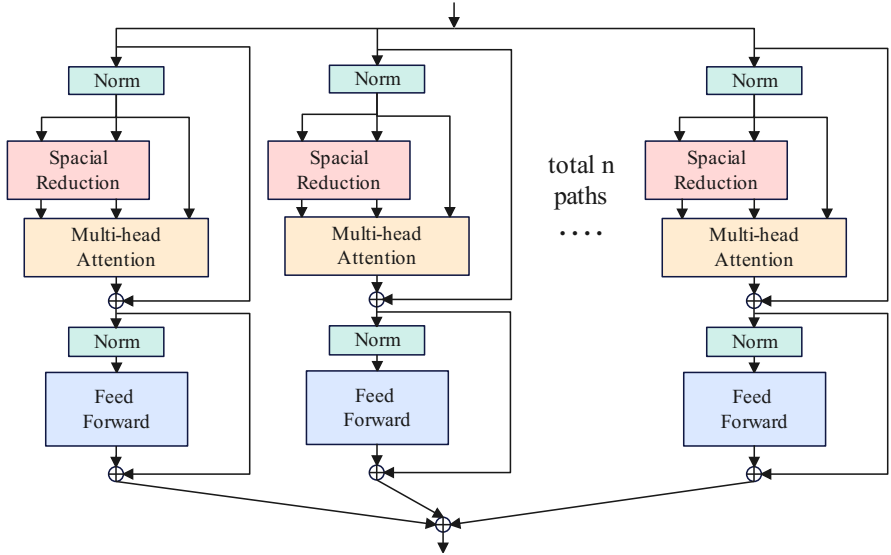

Fig. 4. The architecture of the group encoder. Total *n* paths are *n* group encoders, and *n* are set to 2 and 3, respectively.

## C. Split-transform-merge

The split-transform-merge used by Inception[37] is shown in Fig. 2. In the process of $\sum_{i=1}^{D} w_i x_i$, $x = [x_1, x_2, \cdots, x_D]$ is the input vector of D channels. The strategy splits according to the channel, reshapes each group of channels to obtain a low-dimensional representation, and then transforms and aggregates all the embeddings by addition. Inception uses different transforms for each branch after split, which requires manual design for a specific data set, and has poor generalization ability. And ResNeXt [38] improved on this basis, making all topologies the same. And adding a shortcut, as shown in Eq. (7):

$$F(x) = \sum_{i=1}^{C} \tau_i(x) \quad (6)$$

$$y = x + \sum_{i=1}^{C} \tau_i(x) \quad (7)$$

where Cardinality(C) is the size of the transform set to be aggregated and $\tau_i$ is a sequence of changes.

## III. AGGREGATED PYRAMID VISION TRANSFORMER

### A. Aggregated Transformer

APVT is a pure Transformer framework that introduces a pyramid architecture. The overall architecture is shown in Fig. 3, it consists of multiple improved Transformers as the primary block. Similar to PVT [29], we divide the APVT network architecture into four stages to obtain features of different sizes.

Each Aggregated Transformer (stage) consists of the same parts, including patch embedding and group encoder. They are connected in a similar way to ResNet [43] residual learning, and position embedding [44] is added in the middle.

As shown in Fig. 3, for an input image with the size of Height × Width, the first patch embedding will obtain features of $H/4 \times W/4$, and the number of patches is $H \times W/16$. The next second patch embedding will obtain $1/2 \times 1/2$ features on this basis. For the original input image, the features are $H/8 \times W/8$.

TABLE I. AGGREGATED PYRAMID VISION TRANSFORMER ARCHITECTURE

| Layers | Output Size | APVT-8 | APVT-16 |
|---|---|---|---|
| Patch Embedding (1) | $H/4 \times W/4$ | Patch Size = 4; Channel number = 64 | |
| APVT Encoder (1) | $H/4 \times W/4$ | $\begin{bmatrix} Expansion = 8 \\ Head\ number = 1 \\ Reduction = 8 \end{bmatrix} \times 2$ | $\begin{bmatrix} Expansion = 8 \\ Head\ number = 1 \\ Reduction = 8 \end{bmatrix} \times 3$ |
| Patch Embedding (2) | $H/8 \times W/8$ | Patch Size = 2; Channel number = 128 | |
| APVT Encoder (2) | $H/8 \times W/8$ | $\begin{bmatrix} Expansion = 8 \\ Head\ number = 2 \\ Reduction = 4 \end{bmatrix} \times 2$ | $\begin{bmatrix} Expansion = 8 \\ Head\ number = 2 \\ Reduction = 4 \end{bmatrix} \times 3$ |
| Patch Embedding (3) | $H/16 \times W/16$ | Patch Size = 2; Channel number = 320 | |
| APVT Encoder (3) | $H/16 \times W/16$ | $\begin{bmatrix} Expansion = 4 \\ Head\ number = 5 \\ Reduction = 2 \end{bmatrix} \times 2$ | $\begin{bmatrix} Expansion = 4 \\ Head\ number = 5 \\ Reduction = 2 \end{bmatrix} \times 6$ |
| Patch Embedding (4) | $H/32 \times W/32$ | Patch Size = 2; Channel number = 512 | |
| APVT Encoder (4) | $H/32 \times W/32$ | $\begin{bmatrix} Expansion = 4 \\ Head\ number = 8 \\ Reduction = 1 \end{bmatrix} \times 2$ | $\begin{bmatrix} Expansion = 4 \\ Head\ number = 8 \\ Reduction = 1 \end{bmatrix} \times 3$ |

## B. Group Encoder

According to the split-transform-merge strategy, we can divide the traditional encoder into multiple identical (group) encoders, and then perform operations on each group encoder to operate more efficient feature extraction on the input image, and finally fuse all the group encoders. This work attempts to split the encoder of ViT into multiple group encoders. The splitting and fusion methods are shown in Fig. 4. The aggregated Transformer is composed of group encoders as the primary block, which not only extracts more features, but also improves the accuracy, without adding too much size to the network model.

## C. Feed-Forward

Both the traditional encoder and our newly proposed group encoder are composed of two modules: Multi-Head Attention Layer (MHA) and Feed-Forward Layer. The Feed-Forward Layer maps the MHA-processed vectors to a larger space to facilitate information extraction. The traditional Feed-Forward Layer often uses a fixed-size positional encoding.

Recently, works [45-46] have shown that the fixed-size positional encoding is not friendly to mapping and will lead to lower training accuracy. As shown in Fig. 5, we refer to PVT v2[30] to add a 3 × 3 depthwise convolution to the traditional positional encoding [20] and place it in front of GELU [47] to solve the mapping problem.

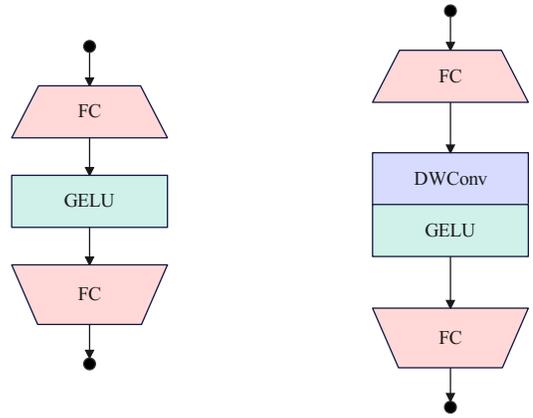

Fig. 5. **Left**: Original Feed-Forward. **Right**: Convolutional Feed-Forward.

## D. Model Details

As shown in TABLE I, we propose two versions of APVT, they are APVT-8 and APVT-16. Taking the 224 × 224 input image used by ImageNet Dataset as an example, both versions of the network architecture will perform 4 patch embeddings and 4 APVT encoders. In order to better compare the experimental results on image classification and object detection tasks, TABLE II details APVT models of different sizes. we set the Heads to Emb values to 32 and 64 respectively, corresponding to the number of heads to determine the embedding size. For example, the embedding sizes of APVT-8-2x-a are 32, 64, 160 and 256, respectively. In addition, we set $n$ paths to 2 and 3, respectively. Different models may use 2 or 3 group encoders.

TABLE II. DETAILS OF AGGREGATED PYRAMID VISION TRANSFORMER

| Model | Layers | Paths | Head to Emb | Heads | Params |
|---|---|---|---|---|---|
| APVT-8-2x-a | 2, 2, 2, 2 | 2 | 32 | 1, 2, 5, 8 | 5.52M |
| APVT-8-2x-b | 2, 2, 2, 2 | 2 | 64 | 1, 2, 5, 8 | 22.88M |
| APVT-16-2x-b | 3, 4, 6, 3 | 2 | 64 | 1, 2, 5, 8 | 42.15M |
| APVT-8-4x-a | 2, 2, 2, 2 | 3 | 32 | 1, 2, 5, 8 | 8.16M |
| APVT-16-4x-a | 3, 4, 6, 3 | 3 | 32 | 1, 2, 5, 8 | 15.48M |

TABLE III. IMAGE CLASSIFICATION PERFORMANCE ON CIFAR-10

| Name | #Param (M) | CIFAR-10 Err (%) | Inference Time (ms) |
|---|---|---|---|
| LeViT-256 [48] | 17.89 | 29.48 | 20 |
| BoTNet-50 [49] | 18.81 | 23.03 | 14 |
| CvT-13 [34] | 19.54 | 21.23 | 21 |
| CeiT-S [50] | 21.17 | 31.28 | 19 |
| APVT-8-2x-b | 22.88 | **20.39** | **17** |
| DeepViT-S [52] | 23.66 | 39.71 | 27 |
| PVTv2-B1 [30] | 25.41 | 21.22 | 25 |
| BoTNet-101 [49] | 32.22 | 22.35 | 25 |
| LeViT-384 [48] | 37.63 | 28.03 | 23 |
| APVT-16-2x-b | 42.15 | **19.55** | **28** |
| DeepViT-B [52] | 50.45 | 36.56 | 33 |

Our network results are in **boldface**, surpassing all competing methods in blue.

## IV. EXPERIMENTAL RESULTS

We use a single GPU RXT 3080Ti to test APVT and other Transformer network architectures for image classification, and use a small model APVT-8-2x-a for object detection tasks.

### A. Image Classification

Image classification experiments are performed on the CIFAR-10 dataset, and all models are not pre-trained for fair comparison. During training, we use a batch size of 128, AdamW [53] to optimize the model. And the initial learning rate is set to $5 \times 10^{-4}$ and decreased by 0.1 every 30 epochs. All models are trained on a single GPU RTX 3080Ti for 60 epochs.

In TABLE III, we see that the APVT model outperforms other Transformer network architectures with a similar number of parameters. For example, when APVT-8-2x-b achieves a similar error rate of CvT-13 and PVTv2-B2, the inference time is reduced to 17ms. In the comparison of large-size models, APVT-16-2x-b achieved the best results with an error rate of 20.58.

TABLE IV. OBJECT DETECTION PERFORMANCE ON COCO val2017

| Method | | AP | $AP_{50}$ | $AP_{75}$ | $AP_S$ | $AP_M$ | $AP_L$ |
|---|---|---|---|---|---|---|---|
| RetinaNet | 6 | 10.4 | 18.0 | 10.6 | 5.3 | 11.4 | |
| | 12 | 16.8 | 27.2 | 17.6 | 8.7 | 17.9 | |

The model is trained for 6/12 epochs using the RetinaNet method, and the backbone is APVT-8-2x-a. The results of object detection are not pre-trained in advance.

### B. Object Detection

Object detection experiments are performed on the COCO 2017 dataset, and we test the performance of APVT-8-2x-a using two standard detectors, RetinaNet [54] and Mask R-CNN [55]. We train with batch size of 16 and optimize by AdamW [53] with an initial learning rate set to $1 \times 10^{-4}$. Our model is trained for 12 epochs on a single GPU RTX 3080Ti.

As shown in TABLE IV, we recorded the Average Precision (AP) values at 6 epcohs and 12 epochs, which were 10.4 and 16.8, respectively. For a small model APVT-8-2x-a with 5.52M parameters, it can obtain 16.8 AP without pre-training, which has excellent performance.

## V. CONCLUSION

In recent years, Transformer has achieved success in computer vision. Many convolutional neural networks have added Transformer architecture to improve model efficiency, and there are also many networks that abandon convolutional layers and use Transformer architecture as the primary block. Compared with the traditional encoder, our proposed group encoder can obtain more features with similar complexity, and the aggregated Transformer is also more efficient than the traditional Transformer. We hope that in future research, there will be more works using the aggregated Transformer to improve the accuracy, not only achieving good results in image classification, but also in tasks such as object detection and semantic segmentation.